\documentclass{article}

\usepackage[final]{nips_2019}

\usepackage[utf8]{inputenc} 
\usepackage[T1]{fontenc}    
\usepackage{hyperref}       
\usepackage{url}            
\usepackage{booktabs}       
\usepackage{amsfonts}       
\usepackage{nicefrac}       
\usepackage{microtype}      
\usepackage{amsmath}
\usepackage{bm}
\usepackage{graphicx}
\usepackage{xcolor}
\usepackage{paralist}
\usepackage{enumitem} 
\usepackage{algorithm}
\usepackage{algorithmic}
\usepackage{subcaption}
\usepackage{multicol,multirow}
\usepackage{wrapfig}
\usepackage[normalem]{ulem}

\usepackage{amssymb}
\DeclareMathOperator*{\softmax}{\text{softmax}}

\DeclareMathOperator*{\argmin}{arg\!\ min}
\DeclareMathOperator*{\argmax}{arg\!\ max}

\newcommand{\tabincell}[2]{\begin{tabular}{@{}#1@{}}#2\end{tabular}}
\newcommand{\secref}[1]{\S\ref{#1}}

\usepackage{cleveref}
\crefformat{section}{\S#2#1#3} 
\crefformat{subsection}{\S#2#1#3}
\crefformat{subsubsection}{\S#2#1#3}

\title{Levenshtein Transformer}

\def \nyu{$^\ddag$}
\def \ti{$^\diamond$}
\def \fair{$^\dagger$}

\author{
Jiatao Gu\fair, 
Changhan Wang\fair,
and
Jake Zhao (Junbo)\nyu\ti\\
\fair Facebook AI Research\\
\nyu New York University \ \ti Tigerobo Inc. \\
\fair\texttt{\{jgu, changhan\}@fb.com} \nyu\texttt{jakezhao@cs.nyu.edu} \\
}

\begin{document}

\maketitle

\begin{abstract}
  Modern neural sequence generation models are built to either generate tokens step-by-step from scratch or (iteratively) modify a sequence of tokens bounded by a fixed length.   In this work, we develop Levenshtein Transformer, a new partially autoregressive model devised for more flexible and amenable sequence generation. Unlike previous approaches, the basic operations of our model are \emph{insertion} and \emph{deletion}.  The combination of them facilitates not only generation but also sequence refinement allowing dynamic length changes.  We also propose a set of new training techniques dedicated at them, effectively exploiting one as the other's learning signal thanks to their complementary nature.  Experiments applying the proposed model achieve  comparable or even better performance with much-improved efficiency on both generation (e.g. machine translation, text summarization) and refinement tasks (e.g. automatic post-editing).  We further confirm the flexibility of our model by showing a Levenshtein Transformer trained by machine translation can straightforwardly be used for automatic post-editing. \footnote{Codes for reproducing this paper are released in \url{https://github.com/pytorch/fairseq/tree/master/examples/nonautoregressive_translation}}

\end{abstract}

\section{Introduction}
Neural sequence generation models are widely developed and deployed in tasks such as machine translation~\citep{bahdanau2014neural,vaswani2017attention}. As we examine the current frameworks, the most popular autoregressive models generate tokens step-by-step.
If not better, recent non-autoregressive approaches~\citep{gu2017non,kaiser2018fast,lee2018deterministic} have proved it possible to perform generation within a much smaller number of decoding iterations.

In this paper, we propose Levenshtein Transformer (LevT), aiming to address the lack of flexibility of the current decoding models.
Notably, in the existing frameworks, the length of generated sequences is either fixed or monotonically increased as the decoding proceeds.
This remains incompatible with human-level intelligence where humans can revise, replace, revoke or delete any part of their generated text.
Hence, LevT is proposed to bridge this gap by breaking the in-so-far standardized decoding mechanism and replacing it with two basic operations --- \emph{insertion} and \emph{deletion}.

We train the LevT using imitation learning. The resulted model contains two policies and they are executed in an alternate manner. Empirically, we show that LevT achieves comparable or better results than a standard Transformer model on machine translation and summarization, while maintaining the efficiency advantages benefited from parallel decoding similarly to ~\citep{lee2018deterministic}.
With this model, we argue that the decoding becomes more flexible. For example, when the decoder is given an empty token, it falls back to a normal sequence generation model.
On the other hand, the decoder acts as a refinement model when the initial state is a low-quality generated sequence. 
Indeed, we show that a LevT trained from machine translation is directly applicable to translation post-editing without any change. This would not be possible with any framework in the literature because generation and refinement are treated as two different tasks due to the model's inductive bias.

One crucial component in LevT framework is the learning algorithm. 
We leverage the characteristics of \emph{insertion} and \emph{deletion} --- they are complementary but also adversarial.
The algorithm we propose is called ``dual policy learning''.
The idea is that when training one policy (insertion or deletion), we use the output from its adversary at the previous iteration as input. 
An \emph{expert} policy, on the other hand, is drawn to provide a correction signal.
Despite that, in theory, this learning algorithm is applicable to other imitation learning scenarios where a dual adversarial policy exists, in this work we primarily focus on a proof-of-concept of this algorithm landing at training the proposed LevT model. 

To this end, we summarize the contributions as follows:
\begin{itemize}[leftmargin=*]
    \item We propose Levenshtein Transformer (LevT), a new sequence generation model composed of the insertion and deletion operations.
    This model achieves comparable or even better results than a strong Transformer baseline in both machine translation and text summarization, but with much better efficiency (up to $\times 5$ speed-up in terms of actual machine execution time);
    \item  We propose a corresponding learning algorithm under the theoretical framework of imitation learning, tackling the complementary and adversarial nature of the dual policies;
    \item We recognize our model as a pioneer attempt to unify sequence generation and refinement, thanks to its built-in flexibility. With this unification, we empirically validate the feasibility of applying a LevT model trained by machine translation directly to translation post-editing, without any change. 
\end{itemize}

\section{Problem Formulation}
\label{sec:formulation}
\subsection{Sequence Generation and Refinement}
We unify the general problems of sequence generation and refinement by casting them to a Markov Decision Process (MDP) defined by a tuple $\left(\mathcal{Y}, \mathcal{A}, \mathcal{E}, \mathcal{R}, \bm{y_0}\right)$.
We consider the setup consisting an agent interacting with an environment $\mathcal{E}$ which receives the agent's editing actions and returns the modified sequence. We define $\mathcal{Y} = \mathcal{V}^{N_{\max}}$ as a set of discrete sequences up to length $N_{\max}$ where $\mathcal{V}$ is a vocabulary of symbols. 
At every decoding iteration, the agent receives an input $\bm{y}$ drawn from scratch or uncompleted generation, chooses an action $\bm{a}$ and gets a reward $r$. We use $\mathcal{A}$ to denote the set of actions and $\mathcal{R}$ for the reward function. 
Generally the reward function $\mathcal{R}$ measures the distance between the generation and the ground-truth sequence, $\mathcal{R}(\bm{y}) = -\mathcal{D}(\bm{y}, \bm{y}^*)$ which can be any distance measurement such as the Levenshtein distance~\citep{levenshtein1965binary}.
It is crucial to incorporate $\bm{y_0} \in \mathcal{Y}$ into the our formulation. As the initial sequence, the agent receives---when $\bm{y_0}$ is an already generated sequence from another system, the agent essentially learns to do refinement while it falls back to generation if $\bm{y_0}$ is an empty sequence.
The agent is modeled by a policy, $\pi$, that maps the current generation over a probability distribution over $\mathcal{A}$. That is,  $\pi : \mathcal{Y} \to P(\mathcal{A})$. 

\subsection{Actions: Deletion \& Insertion}
Following the above MDP formulation, with a subsequence $\bm{y}^k=(y_1, y_2, ..., y_n)$, the two basic actions -- \textit{deletion} and \textit{insertion} -- are called to generate $\bm{y}^{k+1} = \mathcal{E}(\bm{y}^k, \bm{a}^{k+1})$.
Here we let $y_1$ and $y_n$  be special symbols $\texttt{<s>}$ and $\texttt{</s>}$, respectively.
Since we mainly focus on the policy of a single round generation, the superscripts are omitted in this section for simplicity. 
For conditional generation like MT, our policy also includes an input of source information $\bm{x}$ which is also omitted here.

\paragraph{Deletion}
The deletion policy reads the input sequence $\bm{y}$, and 
for every token $y_i \in \bm{y}$, the deletion policy $\pi^{\textrm{del}}(d|i, \bm{y})$ makes
a binary decision which is 1 (delete this token) or 0 (keep it). 
We additionally constrain $\pi^{\textrm{del}}(0|1, \bm{y}) = \pi^{\textrm{del}}(0|n, \bm{y})= 1$ to avoid sequence boundary being broken.
The deletion classifier can also be seen as a fine-grained discriminator used in GAN~\citep{goodfellow2014generative} where we predict ``fake'' or ``real'' labels for every predicted token. 
\paragraph{Insertion} In this work, it is slightly more complex to build the insertion atomic because it involves two phases: \emph{placeholder} prediction and \emph{token} prediction so that it is able to insert multiple tokens at the same slot.
First, among all the possible inserted slots ($y_i, y_{i+1}$) in $\bm{y}$, $\pi^{\textrm{plh}}(p|i, \bm{y})$ predicts the possibility of adding one or several placeholders.
In what follows, for every placeholder predicted as above, a token prediction policy $\pi^{\textrm{tok}}(t | i, \bm{y})$ 
replaces the placeholders with actual tokens in the vocabulary.
The two-stage insertion process can also be viewed as a hybrid of Insertion Transformer~\citep{stern2019insertion} and masked language model~\citep[MLM,][]{devlin2018bert,levy2019constant}.


\paragraph{Policy combination}
Recall that our two operations are complementary. Hence we combine them in an alternate fashion. For example in sequence generation from the empty, insertion policy is first called and it is followed by deletion, and then repeat till the certain stopping condition is fulfilled.
Indeed, it is possible to leverage the parallelism in this combination. We essentially decompose one iteration of our sequence generator into three phases: ``delete tokens -- insert placeholders -- replace placeholders with new tokens''. Within each stage, all operations are performed in parallel. More precisely, given the current sequence $\bm{y}=(y_0, \ldots, y_{n})$, and suppose the action to predict is $\bm{a} = \{\underbrace{d_0,\ldots d_{n}}_{\bm{d}}; \underbrace{p_0, \ldots, p_{n-1}}_{\bm{p}}; \underbrace{t_0^1, \ldots t_0^{p_0}, \ldots, t_{n-1}^{p_{n-1}}}_{\bm{t}}\}$, the policy for one iteration is:
\begin{equation}
    \pi(\bm{a}| \bm{y}) = \prod_{d_i\in \bm{d}} \pi^\textrm{del}(d_i |i, \bm{y})\cdot \prod_{p_i \in \bm{p}} \pi^\textrm{plh}(p_i |i, \bm{y}') \cdot \prod_{t_i \in \bm{t}} \pi^\textrm{tok}(t_i | i, \bm{y}''),
\end{equation}
where $\bm{y}'=\mathcal{E}(\bm{y}, \bm{d})$ and $\bm{y}''=\mathcal{E}(\bm{y}', \bm{p})$. We parallelize the computation within each sub-tasks.


\section{Levenshtein Transformer}
In this section, we cover the specs of Levenshtein Transformer and the dual-policy learning algorithm.
Overall our model takes a sequence of tokens (or none) as the input then iteratively \emph{modify} it by alternating between insertion and deletion, until the two policies combined converge. We describe the detailed learning and inference algorithms in the Appendix.


\begin{figure}
    \centering
    \includegraphics[width=\textwidth]{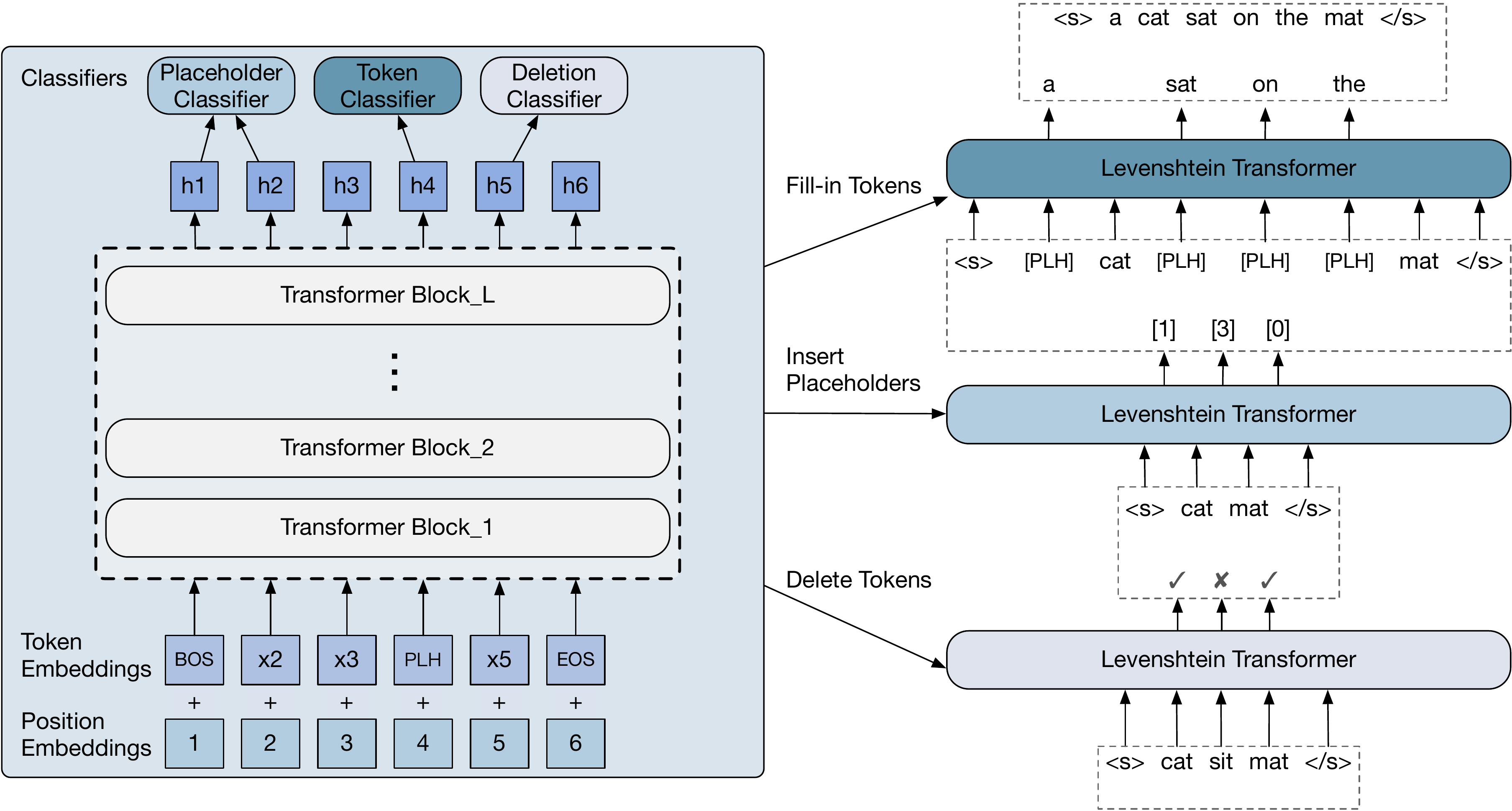}
    \caption{The illustration of the proposed Levenshtein Transformer decoder for \textbf{one refinement iteration}. The same architecture can be applied for three different tasks with specific classifiers. For simplicity, the encoder-decoder attention is omitted within each Transformer-Block.}
    \label{fig:process}
\end{figure}

\subsection{Model}
\label{sec:model}
We use Transformer~\citep{vaswani2017attention} as the basic building block. For conditional generation, the source $\bm{x}$ is included in each TransformerBlock. The states from the $l$-th block are:
\begin{equation}
    \bm{h}_0^{(l+1)}, \bm{h}_1^{(l+1)}, ..., \bm{h}_n^{(l+1)} = \left\{\begin{matrix}
 E_{y_0} + P_0, E_{y_1} + P_1,..., E_{y_n} + P_n,  & l = 0  \\ 
 \text{TransformerBlock}_l(\bm{h}_0^{(l)}, \bm{h}_1^{(l)}, ..., \bm{h}_n^{(l)}), & l > 0
\end{matrix}\right.
\end{equation}
where $E \in \mathbb{R}^{|\mathcal{V}|\times d_\text{model}}$ and $P \in \mathbb{R}^{N_{\text{max}}\times d_\text{model}}$ are the token and position embeddings, respectively. We show the illustration of the proposed LevT model for one refinement (delete, insert) as Figure~\ref{fig:process}. 

\paragraph{Policy Classifiers} The decoder outputs ($\bm{h}_0, \bm{h}_2, ..., \bm{h}_n$) are passed to three policy classifiers:

\begin{enumerate}[leftmargin=*]
    \item \textit{Deletion Classifier}: LevT scans over the input tokens (except for the boundaries) and predict ``deleted'' ($0$) or ``kept'' ($1$) for each token position, 
    \begin{equation}
        \pi_\theta^{\mathrm{del}}(d | i, \bm{y}) = \softmax\left(\bm{h}_i \cdot A^\top\right), \ \ i = 1, \ldots n-1,
        \label{eq.deletion_classifier}
    \end{equation}
    where $A \in \mathbb{R}^{2\times d_\text{model}}$, and we always keep the boundary tokens.
    \item \textit{Placeholder Classifier}: LevT predicts the number of tokens to be inserted at every consecutive position pairs, by casting the representation to a categorical distribution:
        \begin{equation}
            \pi_\theta^{\mathrm{plh}}(p | i, \bm{y}) = \softmax\left(\texttt{concat}(\bm{h}_i, \bm{h}    _{i+1}) \cdot B^\top\right), \ \ i = 0, \ldots n-1,
            \label{eq.placeholder_classifier}
        \end{equation}
    where $B \in \mathbb{R}^{(K_{\max}+1)\times (2d_\text{model})}$. Based on the number ($0\sim K_{\max}$) of tokens it predicts, we insert the considered number of placeholders at the current position. In our implementation, placehoder is represented by a special token $\texttt{<PLH>}$ which was reserved in the vocabulary.
    \item \textit{Token Classifier}: following the placeholder prediction, LevT needs to fill in tokens replacing all the placeholders. This is achieved by training a token predictor as follow:
        \begin{equation}
        \pi_\theta^{\mathrm{tok}}(t | i, \bm{y}) = \softmax\left(\bm{h}_i \cdot C^\top\right), \ \ \forall y_i = \texttt{<PLH>},
            \label{eq.token_classifier}
    \end{equation}
    where $C \in \mathbb{R}^{|\mathcal{V}|\times d_\text{model}}$ with parameters being shared with the embedding matrix.
\end{enumerate}

\paragraph{Weight Sharing} Our default implementation always assumes the three operations to share the same Transformer backbone to benefit features learned from other operations. However, it is also possible to disable weight sharing and train separate decoders for each operations, which increases the capacity of the model while does not affect the overall inference time.
\paragraph{Early Exit} Although it is parameter-efficient to share the same Transformer architecture across the above three heads, there is room for improvement as one decoding iteration requires three full passes of the network. To make trade-off between performance and computational cost, we propose to perform \textit{early exit} (attaching the classifier to an intermediate block instead of the last one) for $\pi^\textrm{del}$ and $\pi^\textrm{plh}$ to reduce computation while keeping $\pi^\textrm{tok}$ always based on the last block, considering that token prediction is usually more challenging than the other two tasks.

\subsection{Dual-policy Learning}
\label{sec.learning}
\paragraph{Imitation Learning} 
We use imitation learning to train the Levenshtein Transformer. Essentially we let the agent imitate the behaviors that we draw from some expert policy $\pi^*$. The expert policy is derived from direct usage of ground-truth targets or less noisy version filtered by sequence distillation~\citep{kim2016sequence}. The objective is to maximize the following expectation:
\begin{equation*}
\begin{split}
      \underbrace{
        \mathbb{E}_{\small\begin{subarray}{l}\bm{y}_\textrm{del} \sim d_{\tilde{\pi}_\textrm{del}}\\\bm{d}^*\sim \pi^*\end{subarray}}\!\!\!
        \sum_{d^*_i \in \bm{d}^*}\!\log \pi_\theta^{\mathrm{del}}(d^*_i|i, \bm{y}_\textrm{del})}_{\textit{Deletion Objective}} + 
      \underbrace{
      \mathbb{E}_{\small\begin{subarray}{l}\bm{y}_\textrm{ins} \sim d_{\tilde{\pi}_\textrm{ins}}\\\bm{p}^*, \bm{t}^*\sim \pi^*\end{subarray}}\!\!
        \left[\sum_{p^*_i \in \bm{p}^*}\!\log \pi_\theta^{\mathrm{plh}}(p^*_i |i, \bm{y}_\textrm{ins}) + \!\!\!
        \sum_{t^*_i \in \bm{t}^*}\!\log \pi_\theta^{\mathrm{tok}}(t^*_i |i, \bm{y}'_\textrm{ins})\right]}_{\textit{Insertion Objective}},
\end{split}
\end{equation*}
where $\bm{y}'_\textrm{ins}$ is the output after inserting palceholders $\bm{p}^*$ upon  $\bm{y}_\textrm{ins}$.
$\tilde{\pi}_\textrm{del}$, $\tilde{\pi}_\textrm{ins}$ are the \textit{roll-in} polices and we repeatedly draw states (sequences) from their induced state distribution $d_{\tilde{\pi}_\textrm{del}}, d_{\tilde{\pi}_\textrm{ins}}$. These states are first executed by the \textit{expert} policy returning the suggested actions by the expert, and then we maximize the conditional log-likelihood over them.
\begin{figure}[htpb]
    \centering
    \includegraphics[width=0.6\textwidth]{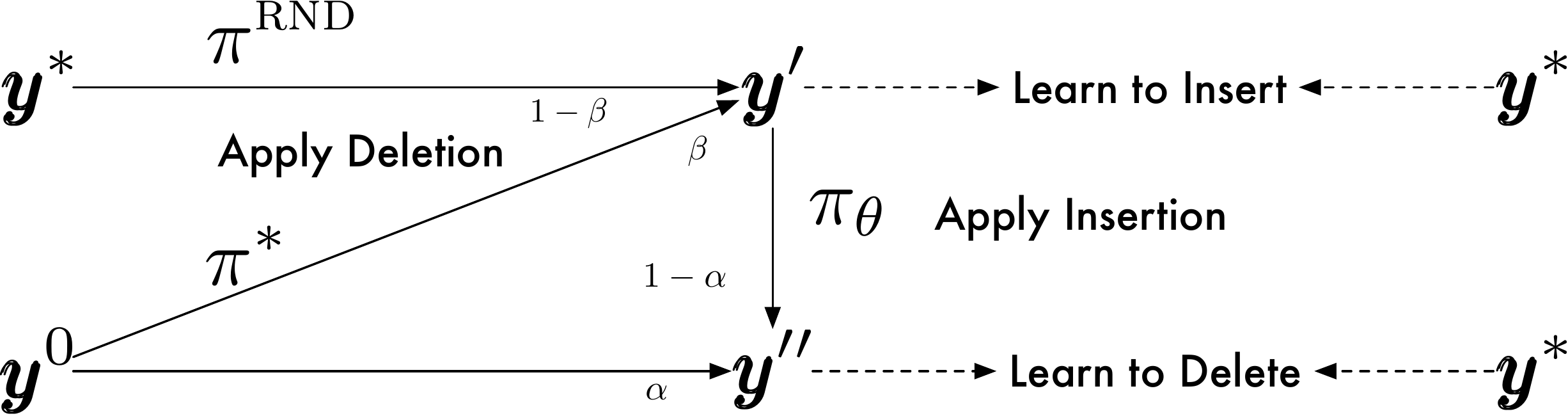}
    \caption{\label{fig.dataflow}The data-flow of learning.}
\end{figure}
By definition, the \textit{roll-in} policy determines the state distribution fed to $\pi_\theta$ during training.
In this work, we have two strategies to construct the roll-in policy --- adding noise to the ground-truth or using the output from the adversary policy. Figure~\ref{fig.dataflow} shows a diagram of this learning paradigm.
We formally write down the roll-in policies as follows.



\begin{enumerate}[leftmargin=*]
    \item \textit{Learning to Delete}: we design the $\tilde{\pi}_\textrm{del}$ as a stochastic mixture between the initial input $\bm{y}^0$ or the output by applying insertion from the model with some mixture factor $\alpha \in [0, 1]$:
        \begin{equation}
            d_{\tilde{\pi}_\textrm{del}} = \{
            \bm{y}^0 \ \ \ \text{if} \ \ u  < \alpha  \ \ \text{else} \ \ \ \mathcal{E}\left(\mathcal{E}\left(\bm{y}', \bm{p}^*\right), \tilde{\bm{t}}\right), \ \ \bm{p}^* \sim \pi^*, \tilde{\bm{t}} \sim \pi_\theta
            \}
        \label{eq.learn.delete}
        \end{equation}
        where $u \sim \text{Uniform}[0, 1]$ and $\bm{y}'$ is any sequence ready to insert tokens. 
        $\tilde{\bm{t}}$ is obtained by sampling instead of doing argmax from Eq.~\eqref{eq.token_classifier}.
    \item \textit{Learning to Insert}: similar to the deletion step, we apply a mixture of the deletion output and a random word dropping sequence of the round-truth, inspired by recent advances of training masked language model~\citep{devlin2018bert}.
    We use random dropping as a form of noise injection to encourage more exploration. 
    Let $\beta \in [0, 1]$ and $u \sim \text{Uniform}[0, 1]$,
        \begin{equation}
        d_{\tilde{\pi}_\textrm{ins}} = \{
            \mathcal{E}\left(\bm{y}^0, \bm{d}^*\right), \ \ \bm{d}^* \sim \pi^* \ \ \ \text{if} \ \ u  < \beta  \ \ \text{else} \
            \ \ \mathcal{E}\left(\bm{y}^*, \tilde{\bm{d}} \right), \ \ \tilde{\bm{d}} \sim \pi^\textsc{rnd}
        \}
        \label{eq.learn.insert}
        \end{equation}
        
\end{enumerate}




\paragraph{Expert Policy}
It is crucial to construct an expert policy in imitation learning which cannot be too hard or too weak to learn from. Specifically, we considered two types of experts:
\begin{enumerate}[leftmargin=*]
    \item \textit{Oracle}: One way is to build an oracle which accesses to the ground-truth sequence. It returns the optimal actions $\bm{a}^*$ (either oracle insertion $\bm{p}^*, \bm{t}^*$ or oracle deletion $\bm{d}^*$) by:
    \begin{equation}
        \bm{a}^* = \argmin_{\bm{a}} \mathcal{D}(\bm{y}^*, \mathcal{E}(\bm{y}, \bm{a}))
    \label{eq.oracle}
    \end{equation}
    Here, we use the Levenshtein distance~\citep{levenshtein1965binary}\footnote{We only consider the variant which only computes insertion and deletion. No substitution is considered.} as $\mathcal{D}$ considering it is possible to obtain the action suggestions efficiently by dynamic programming. 
    \item \textit{Distillation}: 
    We also explore to use another teacher model to provide expert policy, which is known as sequence-level knowledge distillation~\citep{kim2016sequence}.
    This technique has been widely used in previous approaches for nonauoregressive generation~\citep{gu2017non}.
    More precisely, we first train an autoregressive teacher model using the same datasets and then replace the ground-truth sequence $\bm{y^*}$ by the beam-search result of this teacher-model, $\bm{y^\textsc{ar}}$. We use the same mechanism to find the suggested option as using the ground-truth oracle.

    
\end{enumerate}

\subsection{Inference}
\paragraph{Greedy Decoding} 
At inference time, we apply the trained model 
over the initial sequence $\bm{y}^0$ 
for several iterations. We greedily pick up the actions associated with high probabilities in  Eq.~\eqref{eq.deletion_classifier}\eqref{eq.placeholder_classifier}\eqref{eq.token_classifier}. 
Moreover, we find that using search (instead of greedy decoding) or nosiy parallel decoding~\citep{cho2016noisy} does not yield much gain in LevT. 
This observation is quite opposite to what has been widely discovered in autoregressive decoding.
We hypothesize there may be two reasons leading to this issue: (i) The local optimal point brought by greedy decoding in autoregressive models is often far from the optimal point globally. Search techniques resolve this issue with tabularization. In our case, however, because LevT inserts or deletes tokens dynamically, it could easily revoke the tokens that are found sub-optimal and re-insert better ones; (ii) the log-probability of LevT is not a good metric to select the best output. However, we do believe to see more improvements if we include an external re-ranker, e.g. an autoregressive teacher model. We leave this discussion in the future work.




\paragraph{Termination Condition}
The decoding stops when one of the following two conditions is fulfilled:
\begin{enumerate}[leftmargin=*]
    \item \textit{Looping}:
    Generation is terminated if two consecutive refinement iterations return the same output which can be 
    (i) there are no words to delete or insert; 
    (ii) the agent gets stuck in an infinite loop: i.e.  the insertion and deletion counter each other and keep looping.
    
    \item \textit{Timeout}: We further set a maximum number of iterations (timeout) to guarantee a constant-time complexity in the worst case~\citep{lee2018deterministic,levy2019constant}.
\end{enumerate}
\paragraph{Penalty for Empty Placeholders} 
Similar to ~\citet{stern2019insertion}, we add a penalty to insert ``empty'' placeholder in decoding. Overly inserting ``empty'' placeholders may result in shorter output. A penalty term $\gamma \in [0, 3]$ is subtracted from the logits of $0$ in Eq.~\eqref{eq.placeholder_classifier}.


\section{Experiments}
We validate the efficiency, effectiveness, and flexibility of Levenshtein Transformer extensively across three different tasks --- machine translation (MT), text summarization (TS) and automatic post-editing (APE) for machine translation, from both generation (\secref{sec.gen}) and refinement (\secref{sec:seq-ref}) perspectives.

\begin{table}[t]
    \centering
    \caption{\label{table.overall} Generation quality (BLEU $\uparrow$, ROUGE-1/2/L $\uparrow$) and latency (ms $\downarrow$) as well as the average number of decoder iterations ($I_\textsc{dec}$) on the standard test sets for LevT and the autoregressive baseline (with both greedy and beam-search outputs). We show the results of LevT trained from both oracle and the autoregressive teacher model.}
    \scalebox{0.97}{
    \begin{tabular}{rll cc cc}
    \toprule
    & \multirow{2}{*}{Dataset} & \multirow{2}{*}{Metric} & \multicolumn{2}{c}{Transformer} & \multicolumn{2}{c}{Levenshtein Transformer} \\
    &&& greedy & beam4 & oracle & distillation \\
    \midrule
    \multirow{6}{*}{Quality $\uparrow$} 
    &\multirow{1}{*}{Ro-En} & BLEU & 31.67 & 32.30 & 33.02 & \textbf{33.26}\\
    &\multirow{1}{*}{En-De} & BLEU & 26.89 & 27.17 & 25.20 & \textbf{27.27} \\
    &\multirow{1}{*}{En-Ja} & BLEU & 42.86 & \textbf{43.68} & 42.36 & 43.17 \\
    &\multirow{3}{*}{Gigaword} & ROUGE-1 & 37.31 & \textbf{37.87} & 36.14 & 37.40\\
    && ROUGE-2 & 18.10 & \textbf{18.92} & 17.14 & 18.33\\
    && ROUGE-L & 34.65 & \textbf{35.13} & 34.34 & 34.51\\
    \midrule
    \multirow{4}{*}{Speed $\downarrow$} 
    &\multirow{1}{*}{Ro-En} & Latency (ms) /$I_\textsc{dec}$ & 326 / 27.1 & 349 / 27.1  & \ \ 97 / 2.19 & \ \ \textbf{90} / \textbf{2.03} \\
    &\multirow{1}{*}{En-De} & Latency (ms) /$I_\textsc{dec}$ & 343 / 28.1 & 369 / 28.1 & 126 / 2.88 &   \ \ \textbf{92} / \textbf{2.05} \\
    &\multirow{1}{*}{En-Ja} & Latency (ms) /$I_\textsc{dec}$ & 261 / 22.6 & 306 / 22.6 & 112 / 2.61 & \textbf{106} / \textbf{1.97} \\
    &\multirow{1}{*}{Gigaword} & Latency (ms) /$I_\textsc{dec}$ & 116 / 10.1 & 149 / 10.1 &  \ \ 98 / 2.32 & \ \  \textbf{84} / \textbf{1.73} \\
     \bottomrule
    \end{tabular}}
\end{table}
\begin{figure}[t]
    \centering
    \includegraphics[width=\textwidth]{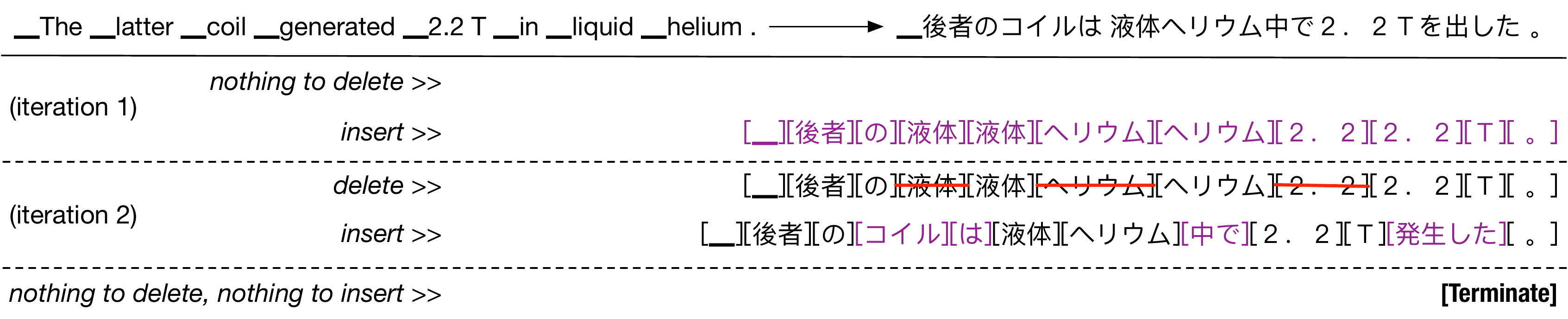}
    \caption{\label{fig.example}An example of WAT'17 En-Ja translation with two decoder iterations by LevT. We present the inserted tokens in {\color{purple} purple} and deleted tokens with {\color{red} \sout{red strikethrough}}}.
    \vspace{-5pt}
\end{figure}
\subsection{Sequence Generation}
\label{sec.gen}
For the sequence generation perspective, we evaluate LevT model on MT and TS. As a special case, sequence generation assumes empty $\bm{y}^0=\textsc{<s></s>}$ as input and no initial deletion is applied. 

\paragraph{Data \& Evaluation} 
We use three diversified language pairs for MT experiments: WMT'16 Romanian-English (Ro-En)\footnote{\url{http://www.statmt.org/wmt16/translation-task.html}}, WMT'14 English-German (En-De)\footnote{\url{http://www.statmt.org/wmt14/translation-task.html}} and WAT2017 Small-NMT English-Japanese~\citep[En-Ja,][]{nakazawa-etal-2017-overview}\footnote{\url{http://lotus.kuee.kyoto-u.ac.jp/WAT/WAT2017/snmt/index.html}}. The TS experiments use preprocessed data from the Annotated English Gigaword~\citep[Gigaword,][]{rush2015neural}\footnote{\url{https://github.com/harvardnlp/sent-summary}}. 
We learn byte-pair encoding~\citep[BPE,][]{sennrich2015neural} vocabulary on tokenized data.
Detailed dataset statistics can be found in the Appendix. For evaluation metrics, we use BLEU~\citep{papineni2002bleu} for MT and ROUGE-1,2,L~\citep{lin:2004:ACLsummarization} for TS. Before computing the BLEU scores for Japanese output, we always segment Japanese words using KyTea~\footnote{\url{http://www.phontron.com/kytea/}}.

\paragraph{Models \& Training} We adopt the model architecture of Transformer base~\citep{vaswani2017attention} for the proposed LevT model and the autoregressive baseline. 
All the Transformer-based models are trained on $8$ Nvidia Volta GPUs with maximum $300K$ steps and a total batch-size of around $65,536$ tokens per step (We leave more details to the Appendix).

\paragraph{Overall results} 
We present our main results on the generation quality and decoding speed in Table~\ref{table.overall}. We measure the speed by the averaged generation latency of generating one sequence at a time on single Nvidia V100 GPU. To remove the implementation bias, we also present the number of decoder iterations as a reference. It can be concluded that for both MT and summarization tasks, our proposed LevT achieves comparable and sometimes better generation quality compared to the strong autoregressive baseline, while LevT is much more efficient at decoding. A translation example is shown in Figure~\ref{fig.example} and we leave more in Appendix.
We conjecture that this is due to that the output of the teacher model possesses fewer modes and much less noisy than the real data. Consequently, LevT needs less number of iterations to converge to this expert policy.

\begin{table}[t]
\centering
    \caption{Ablation study for Levenshtein Transformer on En-De (a) and  Ro-En (b) translation tasks.}
    \scalebox{0.97}{
\begin{subtable}[t]{.66\textwidth}
    \raggedright
    \caption{\label{table.weight-share}Test BLEU for variant weight sharing. 
    Baseline scores from~\citet[IT,][]{lee2018deterministic},~\citet[MaskT,][]{levy2019constant} are included for reference.}
    \begin{tabular}{lcccc|cc}
    \toprule
    sharing & $\textrm{none}$ & $\textrm{plh},\textrm{ins}$ & $\textrm{ins},\textrm{del}$ & $\textrm{all}$ & IT & MaskT \\ 
    \midrule
    \textit{oracle} & $-$ & 25.50 & $-$ & 25.20 & $-$ & $-$ \\ 
    \textit{distill} & 25.11 & \textbf{27.73} & 24.90 & 27.27 & 21.61 & 26.56 \\
    \bottomrule
    \end{tabular}
\end{subtable}
\hspace{5pt}
\begin{subtable}[t]{.32\textwidth}
    \centering
    \caption{\label{table.roll-in} Test BLEU and deletion loss with variant roll-in polices.}
    \begin{tabular}{lrr}
    \toprule
    roll-in & BLEU & $\textrm{NLL}$(del) \\ 
    \midrule
    \textit{Ours} & \textbf{33.02} & $\approx$ 0.202 \\ 
    \textit{DAE} & 31.78 & $\approx$  0.037 \\
    \bottomrule
    \end{tabular}
\end{subtable}}
\end{table}

\begin{figure}[t]
    \centering
    \begin{subfigure}[b]{0.58\textwidth}
    \raggedleft
    \includegraphics[width=\textwidth]{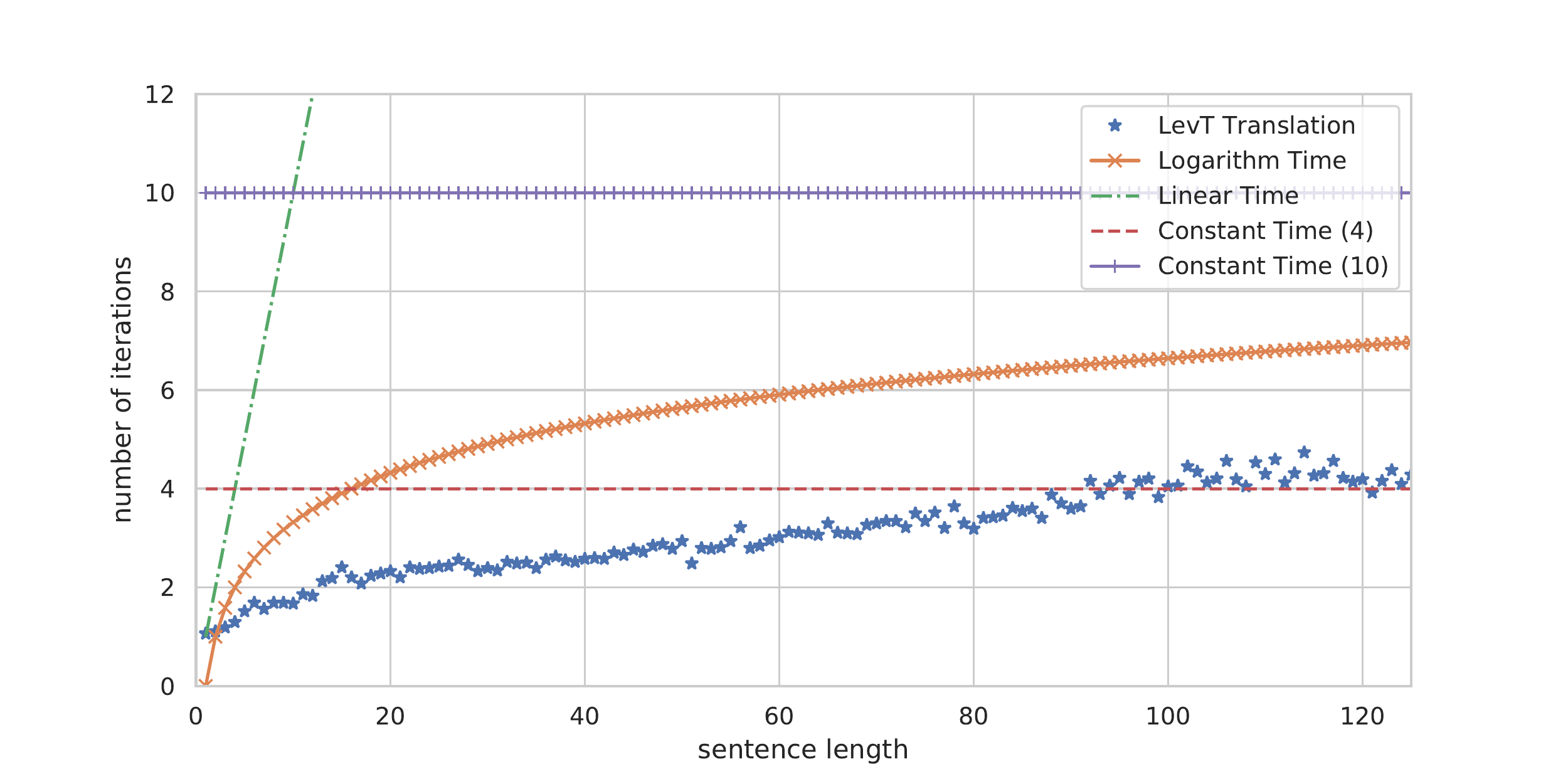}
    \caption{\label{fig.iter_length} Average number of refinement iterations v.s. length measured on monolingual corpus. 
                For most of the time, LevT decodes with much smaller number (generally, 1$\sim$4) of iterations.}
    \end{subfigure}
    \hspace{2.5pt}
    \begin{subfigure}[b]{0.385 \textwidth}
    \raggedright
    \includegraphics[width=\textwidth]{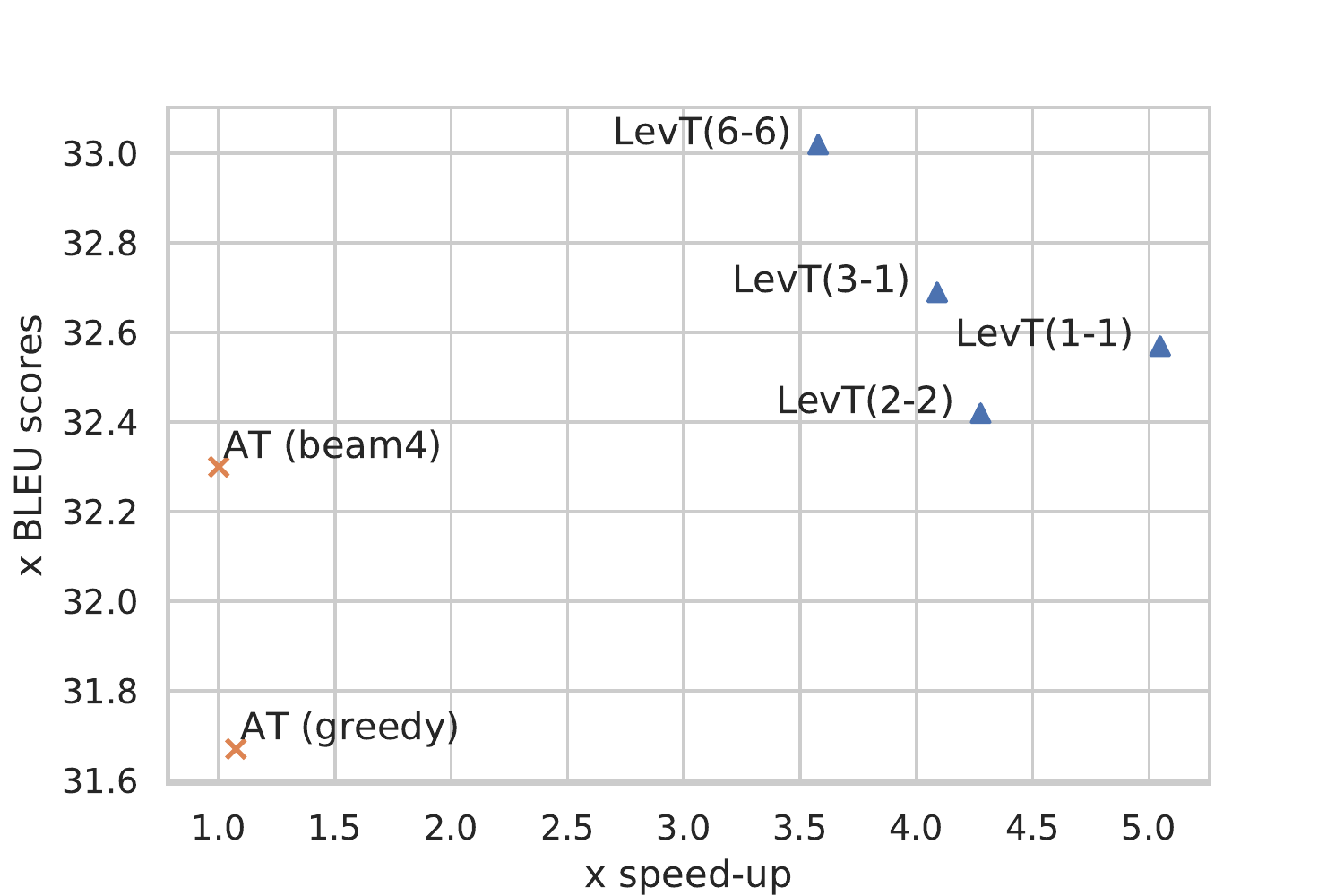}
    \caption{\label{fig.speedup} BLEU v.s. speed-up for LevT across variant early-exits and the autoregressive baselines on the test set of Ro-En. }
    \end{subfigure}
    \caption{\label{fig.efficiency}Plots showing the decoding efficiency of the proposed Levenshtein Transformer.}
    \vspace{-5pt}
\end{figure}
%

\paragraph{Ablation on Efficiency}
As shown in Figure~\ref{fig.iter_length}, we plot the average number of iterations over the length of input over a monolingual corpus. LevT learns to properly adjust the decoding time accordingly. We also explore the variants of ``early exit'' where we denote LevT($m$-$n$) as a model with $m$ and $n$ blocks for deletion (Eq.~\eqref{eq.deletion_classifier}) and placeholder prediction (Eq.~\eqref{eq.placeholder_classifier}) respectively. Figure~\ref{fig.speedup} shows that although it compromises the quality a bit, our model with early exit achieves up to $\times 5$ speed-up (execution time) comparing against a strong autoregressive Transformer using beam-search.

\paragraph{Ablation on Weight Sharing}
We also evaluate LevT with different weight sharing as noted in \secref{sec:model}. The results of models trained with oracle or distillation are listed in Table~\ref{table.weight-share}.
We observe that weight-sharing is beneficial especially between the two insertion operations (placeholder and token classifiers). Also, it shows another $+0.5$ BLEU improvement by not sharing the deletion operation with insertion compared to the default setting, which may indicate that insertion and deletion capture complementary information, requiring larger capacity by learning them separately.

\paragraph{Importance of mixture roll-in policy} 
We perform an ablation study on the learning algorithm. Specifically, we train a model with no mixing of the $\pi_\theta$ in Equation~\eqref{eq.learn.delete}. We name this experiment by $\textrm{DAE}$ due to its resemblance to a denoising autoencoder.
We follow closely a standard pipeline established by~\citet{lee2018deterministic}.
Table~\ref{table.roll-in} shows this comparison. As we can see that the deletion loss from $\textrm{DAE}$ is much smaller while the generation BLEU score is inferior.
We conjecture that this is caused by the mismatch between the states from the model and the roll-in policy in training the $\textrm{DAE}$.

\paragraph{v.s. Exiting Refinement-based Models}
Table~\ref{table.weight-share} also includes results from two relevant recent works which also incorporate iterative refinement in non-autoregressive sequence generation.
For fair comparison,  we use the result with length beam $1$ from \citet{levy2019constant}. Although both approaches use similar ``denosing'' objectives to train the refinement process, our model explicitly learns ``insertion'' and ``deletion'' in a dual-policy learning fashion, and outperforms both models.

\subsection{Sequence Refinement}
We evaluate LevT's capability of refining sequence outputs on the APE task. In this setting, inputs are pairs of the source sequence and a black-box MT system generation. The ground-truth outputs are from real human edits with expansion using synthetic data.

\label{sec:seq-ref}
\paragraph{Dataset}
We follow a normal protocol in the synthetic APE experiments~\citep{grangier2017quickedit}:
we first train the input MT system on half of the dataset. Then we will train a refinement model on the other half based on the output produced by the MT model trained in the previous phase.
For the real APE tasks, we use the data from WMT17 Automatic Post-Editing Shared Task\footnote{\url{http://www.statmt.org/wmt17/ape-task.html}} on En-De. It contains both real PE triples and a large-scale synthetic corpus.
\begin{table}[t]
    \centering
    \caption{\label{table.ape}Performance (BLEU $\uparrow$ / case-sensitive TER $\downarrow$) comparison on APE. ``do nothing'' represents the results of the original MT system output; the autoregressive model uses beam-size $4$. For the proposed LevT, we use ``scratch'' to denote training from scratch on the APE triple data, and use ``zero-shot'' to denote applying an MT pre-trained LevT model directly for post-editing tasks. The same model can be further fine-tuned. 
    All scores with \underline{underlines} are from the model trained with an autoregressive teacher model (distillation) as the expert policy.}
    \scalebox{0.92}{
    \begin{tabular}{lll cc cccc}
    \toprule
    \multicolumn{2}{c}{\multirow{2}{*}{Dataset}} &  
    MT & \multirow{2}{*}{Do-Nothing} & \multirow{2}{*}{Transformer} & \multicolumn{3}{c}{Levenshtein Transformer} \\ 
    & & system & & & Scratch & Zero-shot & Fine-tune & \\
    \midrule
    \multirow{4}{*}{Synthetic} 
    & \multirow{2}{*}{Ro-En}  & PBMT  & 27.5 / 52.6 & 28.9 / 52.8 & 29.1 / \textbf{50.4} &  \textbf{30.1} / 51.7 & $-$ \\
                            & & NMT & 26.2 / 56.5 &  26.9 / 55.6 & \textbf{28.3} / \textbf{53.6} &   28.0 / 55.8 & $-$  \\
    & \multirow{1}{*}{En-De}  & PBMT & 15.4 / 69.4 & 22.8 / 61.0 & \textbf{\underline{25.8}} / \textbf{\underline{56.6}} &   \underline{16.5} / \underline{69.6} & $-$  \\    
    & \multirow{1}{*}{En-Ja}  & NMT &  37.7 /  48.0 & 41.0 / 44.9 & \textbf{\underline{42.2}} / \textbf{\underline{44.3}} &  \underline{39.4} / \underline{47.5} & $-$ \\    
    \midrule
    Real &  En-De & PBMT & 62.5 / 24.5 & 67.2 / 22.1 & 66.9 / 21.9 & 59.6 / 28.7 & \textbf{70.1} / \textbf{19.2} \\
     \bottomrule
    \end{tabular}}

\end{table}
\begin{figure}[t]
    \centering
    \begin{subfigure}[b]{0.4\textwidth}
    \includegraphics[width=\textwidth]{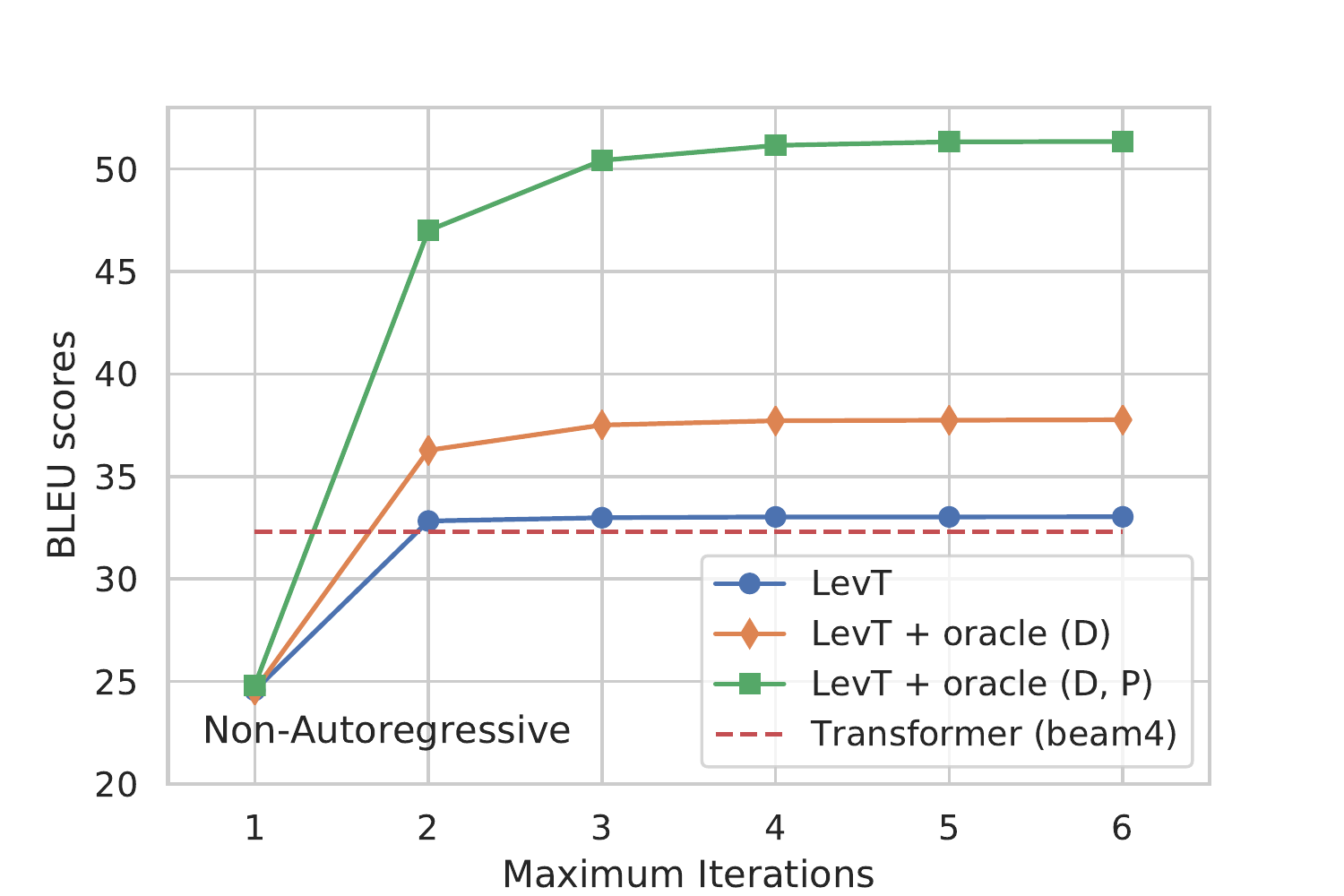}
    \caption{Test set BLEU scores for WMT Ro-En}
    \end{subfigure}
    \hspace{0.12\textwidth}
    \begin{subfigure}[b]{0.4\textwidth}
    \includegraphics[width=\textwidth]{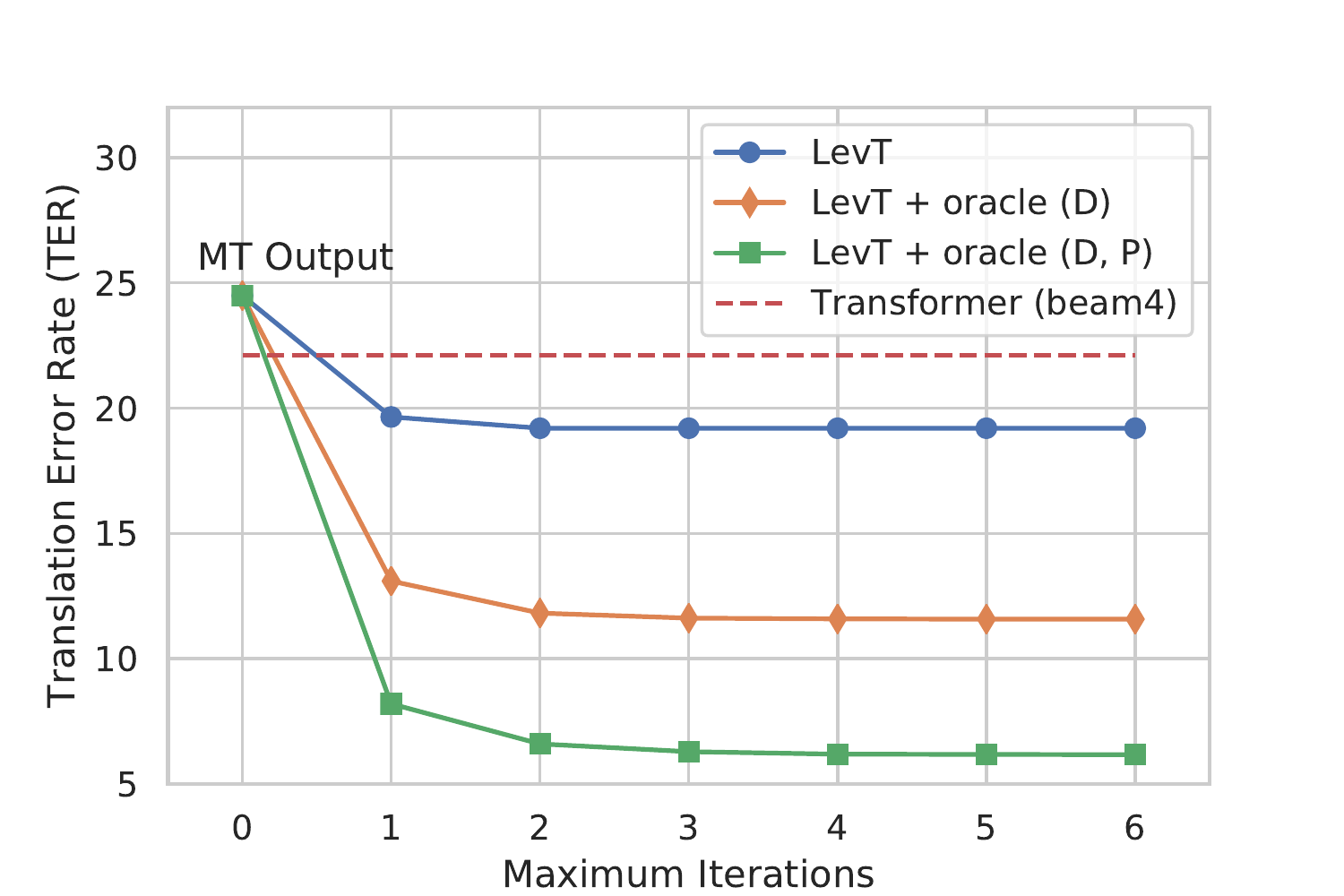}
    \caption{Test set TER scores for Real APE En-De}
    \end{subfigure}
    \caption{\label{fig.oracle} MT \& PE Performance v.s. Timeout iterations w/o oracle instructions.}
\end{figure}
\paragraph{Models \& Evaluation} 
The baseline model is a standard Transformer encoding the concatenation of the source and the MT system's output.
For the MT system here, we want some imperfect systems that need to be refined. 
We consider a statistical phrase-based MT system~\citep[PBMT,][]{koehn2003statistical} 
and an RNN-based NMT system~\citep{bahdanau2014neural}.
Apart from BLEU scores, we additionally apply translation error rate~\citep[TER,][]{snover2006study} as it is widely used in the APE literature.


\paragraph{Overall results}
We show the major comparison in Table~\ref{table.ape}. When training from scratch, LevT consistently improves the performance of the input MT system (either PBMT or NMT). It also achieves better performance than the autoregressive Transformer in most of the cases.
%
%
\paragraph{Pre-training on MT}
Thanks to the generality of the LevT model, we show it is feasible to directly apply the LevT model trained by generation onto refinement tasks --- in this case --- MT and APE. 
We name this a ``zero-shot post-editing'' setting.
According to Table~\ref{table.ape}, the pre-trained MT models are always capable of improving the initial MT input in the synthetic tasks.


The real APE task, however, differs quite a bit from the synthetic tasks because human translators normally only fix a few spotted errors. This ends up with very high BLEU scores even for the ``Do-nothing'' column. However, the pre-trained MT model achieves the best results by fine-tuning on the PE data indicating that LevT is able to leverage the knowledge for generation and refinement.


\paragraph{Collaborate with Oracle}
Thanks to the saperation of \textit{insertion} and \textit{deletion} operations, LevT has better interpretability and controllability. For example, we test the ability that LevT adapts oracle (e.g. human translators) instructions. As shown in Figure~\ref{fig.oracle}, both MT and PE tasks have huge improvement if every step the oracle \textit{deletion} is given. This goes even further if the oracle provides both the correct \textit{deletion} and the number of \textit{placehoders} to insert. It also sheds some light upon computer-assisted text editing for human translators.


\section{Related Work}
\paragraph{Non-Autoregressive and Non-Monotonic Decoding} Breaking the autoregressive constraints and monotonic (left-to-right) decoding order in classic neural sequence generation systems has recently attracted much interest. \cite{stern2018blockwise,wang2018semi} designed partially parallel decoding schemes to output multiple tokens at each step. 
\cite{gu2017non} proposed a non-autoregressive framework using discrete latent variables, which was later adopted in 
\cite{lee2018deterministic} as iterative refinement process.
\cite{levy2019constant} introduced the masked language modeling objective from BERT~\citep{devlin2018bert} to non-autoregressively predict and refine translations.  \cite{welleck2019non,stern2019insertion, gu2019insertion} generate translations non-monotonically by adding words to the left or right of previous ones or by inserting words in arbitrary order to form a sequence. 

\paragraph{Editing-Based Models} 
Several prior works have explored incorporating ``editing'' operations for sequence generation tasks. For instance,
\cite{novak2016iterative} predict and apply token substitutions iteratively on phase-based MT system outputs using convolutional neural network. QuickEdit~\citep{grangier2017quickedit} and deliberation network~\citep{xia2017deliberation} both consist of two autoregressive decoders where the second decoder refines the translation generated by the first decoder. \citet{guu2018generating} propose a neural editor which learned language modeling by first retrieving a prototype and then editing over that. \cite{freitag2019text} correct patterned errors in MT system outputs using transformer models trained on monolingual data.
Additionally, the use of Levenshtein distance with dynamic programming as the oracle policy were also proposed in ~\citet{sabour2018optimal,dong2019editnts}. Different from these work, the proposed model learns a non-autoregressive model which simultaneously inserts and deletes multiple tokens iteratively.

\section{Conclusion}
We propose Levenshtein Transformer, a neural sequence generation model based on insertion and deletion. The resulted model achieves performance and decoding efficiency, and embraces sequence generation to refinement in one model. 
The insertion and deletion operations are arguably more similar to how human writes or edits text. For future work, it is potential to extend this model to human-in-the-loop generation.

\section*{Acknowledgement}
We would like to thank Kyunghyun Cho, Marc'Aurelio Ranzato, Douwe Kiela, Qi Liu and our colleagues at Facebook AI Research for valuable feedback, discussions and technical assistance.

\bibliographystyle{acl_natbib}
\bibliography{nips}
\newpage
\appendix
\section{Learning \& Inference Algorithm}
We present the detailed algorithms for learning and decoding from Levenshtein Transformer as follows. For simplicity, we always omit the source information $\bm{x}$ in conditional sequence generation tasks such as machine translation which is handled by the cross-attention with an encoder on $\bm{x}$.

The learning algorithm is shown in Algorithm~\ref{alg:learning}. $\mathcal{E}$ is the environment and $\mathcal{D}$ is denoted as the Levenshtein distance, and we can easily back-track the optimal insertion and deletion operations
through dynamic programming. We only show the the case with single batch-size for convenience.
We also present the inference algorithm in Algorithm~\ref{alg:decoding}. If the initial sequence $\bm{y}^0$
is empty (\texttt{<s></s>}), the
proposed model will skip the first deletion and do sequence generation. Otherwise, the model starts
with deletion operations and refine the input sequence.

\begin{algorithm}[htpb]
	  \caption{Learning for Levenshtein Transformer}
  		\label{alg:learning}
\begin{algorithmic}
\STATE {\bfseries Initialize:} Training set $\mathcal{T}$, expert policy $\pi^*$, model policy $\pi_\theta$, random deletion policy $\pi^\textsc{rnd}$, $\alpha$, $\beta$
\REPEAT
	\STATE {Sample a training pair $(\bm{y}^0, \bm{y}^*) \sim \mathcal{Y}$}
	\IF {expert $\pi^*$ is a teacher model}
		\STATE Set the teacher's output as the target {$\bm{y}^* = \bm{y}^\textsc{ar}$}
	\ENDIF
	\STATE {Sample $u, v \sim \text{Uniform}[0, 1]$}
	\IF {$u < \beta$}
    	\STATE {$\bm{y}_\textrm{ins} = \mathcal{E}(\bm{y}^0, \bm{\tilde{d}})$, where $\bm{\tilde{d}}=\argmin_{\bm{d}}\mathcal{D}(\bm{y}^*, \mathcal{E}\left(\bm{y}^0, \bm{d})\right)$}
	\ELSE
	    \STATE {$\bm{y}_\textrm{ins} = \mathcal{E}(\bm{y}^*, \bm{\tilde{d}})$, where $\bm{\tilde{d}} \sim \pi^\textsc{rnd}(\cdot | \bm{y}^*)$}
	\ENDIF
	\STATE {$\bm{y}_\textrm{ins}' = \mathcal{E}(\bm{y}_\textrm{ins}, \bm{p}^*)$, where $\bm{p}^*,\bm{t}^*=
	\argmin_{\bm{p},\bm{t}}\mathcal{D}(\bm{y}^*, \mathcal{E}\left(\bm{y}_\textrm{ins}, \{\bm{p}, \bm{t}\})\right)$}
	\IF {$v < \alpha$}
	    \STATE {$\bm{y}_\textrm{del} = \bm{y}^0$}
	\ELSE 
	    \STATE {$\bm{y}_\textrm{del} = \mathcal{E}(\bm{y}_\textrm{ins}', \bm{\hat{t}})$, where $\bm{\hat{t}} = \argmax_{\bm{t}}\sum_{y_i \in \bm{y}_\textrm{ins}', y_i=\texttt{<PLH>}}\log\pi_\theta^\textrm{tok}(t_i | i, \bm{y}_\textrm{ins}')$}
	\ENDIF
	\STATE {$\mathcal{L}^\textrm{ins}_\theta = -\left[
 	\sum_{y_i \in \bm{y}_\textrm{ins}, p^*_i\in\bm{p}^*} 
 	\log \pi_\theta^\textrm{plh}(p^*_i |i, \bm{y}_\textrm{ins}) +
 	\sum_{y_i \in \bm{y}_\textrm{ins}', y_i=\texttt{<PLH>}, t^*_i\in \bm{t}^*} 
 	\log \pi_\theta^\textrm{tok}(t^*_i |i, \bm{y}_\textrm{ins}') \right]$}
	\STATE { $\mathcal{L}^\textrm{del}_\theta = -\sum_{y_i \in \bm{y}_\textrm{del}, d^*_i \in \bm{d}^*} \log \pi_\theta^\textrm{del}(d^*_i |i, \bm{y}_\textrm{del})$, where $\bm{d}^*=\argmin_{\bm{d}}\mathcal{D}(\bm{y}^*, \mathcal{E}\left(\bm{y}_\textrm{del}, \bm{d})\right)$}
    \STATE $\theta = \theta - \lambda \cdot \bigtriangledown_\theta
    \left[\mathcal{L}^\textrm{ins}_\theta + \mathcal{L}^\textrm{del}_\theta\right]$
\UNTIL{Maximum training steps reached}
\end{algorithmic}
\end{algorithm}
\begin{algorithm}[htpb]
  \caption{Decoding for Levenshtein Transformer}
  \label{alg:decoding}
\begin{algorithmic}
  \STATE {\bfseries Initialize:} Input $\bm{y} = \bm{y}^0$, step $t=0$, maximum step $T_{\max}$, model policy $\pi_\theta$.
  \REPEAT
  \IF{$\bm{y} = \texttt{<s>}\texttt{</s>}$}
  	\STATE {Empty sequence, skip deletion: $\bm{y}'=\bm{y}$}
  \ELSE
  	\STATE {Delete tokens: $\bm{y}' = \mathcal{E}(\bm{y}, \bm{\hat{d}})$, where $\bm{\hat{d}} = \argmax_{\bm{d}} \sum_{y_i \in \bm{y}}\log\pi_\theta^\textrm{del}(d_i | i, \bm{y})$}
  \ENDIF
  \IF {(t > 0) \& ($\bm{y}' = \bm{\tilde{y}}$)}
  \STATE {Termination condition satisfied: direct loop}
  \STATE {\textbf{break}}
  \ENDIF
  \STATE {Assign deleted output for back-up $\bm{\tilde{y}} = \bm{y}'$}
  \STATE {Insert placeholders: $\bm{y}'' = \mathcal{E}(\bm{y}', \bm{\hat{p}})$, where $\bm{\hat{p}} = \argmax_{\bm{p}}\sum_{y_iy_{i+1} \in \bm{y}'}\log\pi_\theta^\textrm{plh}(p_i | i, \bm{y}')$}
  \IF {$\bm{y}''=\bm{y}'=\bm{y}$}
  	\STATE {Termination condition satisfied: nothing to delete, nothing to insert.}
  	\STATE {\textbf{break}} 
  \ENDIF
  \IF {$\bm{y}''=\bm{y}'$}
  	\STATE {Nothing to insert, skip insertion: $\bm{y} = \bm{y}''$}
  \ELSE
	\STATE {Replace placeholders: $\bm{y} = \mathcal{E}(\bm{y}'', \bm{\hat{t}})$, where $\bm{\hat{t}} = \argmax_{\bm{t}}\sum_{y_i \in \bm{y}'', y_i=\texttt{<PLH>}}\log\pi_\theta^\textrm{tok}(t_i | i, \bm{y}'')$}
  \ENDIF
  \STATE {Update steps: t = t + 1}
  \UNTIL{Reach the maximum length $t = T_{\max}$}
  \STATE{\textbf{return} $\bm{y}$}
\end{algorithmic}
\end{algorithm}

\section{Dataset and Preprocessing Details}
Table \ref{tab:dataset_stats_gen} and \ref{tab:dataset_stats_refine} list the statistics ($\#$ of sentences, vocabulary) for all the datasets used in this work. 
 We learn BPE vocabulary with $32,000$ joint operations for WMT En-De and Gigaword and $40,000$ joint operations for WMT Ro-En. For WAT En-Ja, we adopt the official $16,384$ BPE vocabularies learned separately on source and target side.
\begin{table}[htpb]
    \centering
    \caption{Dataset statistics for sequence generation tasks (MT and TS).}
    \begin{tabular}{ll|rrrr}
    \toprule
    \multicolumn{2}{c|}{Dataset} &  Train & Valid & Test & Vocabulary\\
    \midrule
    \multirow{3}{*}{Translation} & 
      WMT'16 Ro-En  & 608,319 & 1999 & 1999 & 34,983\\
    & WMT'14 En-De  & 4,500,966 & 3000 & 3003 & 37,009\\
    & WAT'17 \  En-Ja  & 2,000,000 & 1790 & 1812 & 17,952 / 17,801\\
    \midrule
    Summarization & English Gigaword & 3,803,957 & 189,651 & 1951 & 30,004\\
    \bottomrule
    \end{tabular}
    
    \label{tab:dataset_stats_gen}
\end{table}
\begin{table}[htpb]
    \centering
    \caption{Dataset statistics for sequence refinement tasks (APE).}
    \begin{tabular}{ll|rrrrr}
    \toprule
    \multicolumn{2}{c|}{Dataset} & MT-Train & APE-Train & Valid & Test & Vocabulary\\
    \midrule
    \multirow{3}{*}{Synthetic}
    & WMT'16 Ro-En  & 300,000 & 308,319 & 1999 & 1999 & 34,983\\
    & WMT'14 En-De  & 2,250,000 & 2,250,967 & 3000 & 3003 & 37,009 \\
    & WAT'17 \  En-Ja & 1,000,000 & 1,000,000 & 1790 & 1812 & 17,952 / 17,801\\
    \midrule
    Real & \tabincell{c}{WMT'17 APE\\ En-De} & 4,391,180 & \tabincell{c}{526,368 (fake)\\ + 24,000 (real)} & 2000 & 2000 & 40,349 \\
    \bottomrule
    \end{tabular}
    
    \label{tab:dataset_stats_refine}
\end{table}
\section{Model and Training Details}
\subsection{Sequence Generation Tasks}
\label{apx.gen}
Transformer models are used for autoregressive baselines as well as teacher models (for the expert policy). By default, we set $d_\text{model}=512$, $d_\text{hidden}=2048$, $n_\text{heads}=8$, $n_\text{layers}=6$, $\text{lr}_{\max}=0.0005$, $\text{label-smooth}=0.1$, $\text{warmup}=10000$ and $\text{dropout}=0.3$. Source and target side share embeddings in all the training pairs except for WAT En-Ja where BPE vocabularies of both side are learned separately and are almost non-overlapping. 

Since the training objectives for Levenshtein Transformer contains randomness terms (Eq.~\eqref{eq.learn.delete} \eqref{eq.learn.insert}), we instead use BLEU (for MT) or ROUGE-2 (for TS) to select the best checkpoint by validation scores. We do not average checkpoints in this work.

\subsection{Sequence Refinement Tasks}
For synthetic APE tasks, we keep the same training conditions for LevT as those for MT tasks (\secref{apx.gen}). 
As described earlier in \secref{sec:seq-ref}, we build the baseline Transformer by concatenating
the source and MT system’s output as the input sequence for the encoder. Specially, we restart the
positional embeddings for the MT output, add an additional language embedding for each token of the input sequence to show its language type. The detailed hyperpameters are the same as the
standard Transformer.

As described in \secref{sec:seq-ref}, we consider the following two different imperfect MT systems to provide the
refinement inputs. Firstly, we consider the traditional statistical phrase-based machine translation
system (PBMT). We follow the instruction to build the basic baseline model via moses\footnote{\url{http://www.statmt.org/moses/?n=Moses.Baseline}}. 
As for the NMT-based model, we use a single layer attention-based model composed by LSTM. We build this
model on fairseq-py\footnote{\url{https://github.com/pytorch/fairseq/blob/master/fairseq/models/lstm.py}} with the default configuration.

For the real APE task, we follow the procedures introduced in Junczys-Dowmunt and Grundkiewicz
(2016). Synthetic corpus has two subsets: a 500K one and a 4M one. We over-sample real data by 10
times and merge it with the 500K synthetic data to train APE models. Besides, we also train a LevT
MT model on the bigger (4M) synthetic corpus where we only use the source and target pairs.

\subsection{Implementation}
Both the proposed Levenshtein Transformer and the baseline Transformer are implemented using
PyTorch\footnote{\url{https://pytorch.org/}}. The codes are released as part of the Fairseq-py~\footnote{\url{https://github.com/pytorch/fairseq/tree/master/examples/nonautoregressive_translation}}.

\begin{table}[]
    \centering
    \caption{The percentage of WMT En-De test sentence generation terminated at each iteration using LevT(T) with a maximum iteration of $10$.}
    \begin{tabular}{ccccccccccc|c}
\\\toprule  
Iterations & 1 & 2 &3 &4 &5 &6 &7&8 & 9 &10& 2.43\\\midrule
\% & 12.3 & 48.1 & 28.5 & 8.5 & 2.0 & 0.4 & 0.1 & 0 & 0 & 0.1 & AVG
 \\  \bottomrule
\end{tabular}   
    \label{tab:max_iteration}
\end{table}

\subsection{Maximum Number of Iterations}
We also presented in general how many sentences will be generated using the maximum iteration (for instance $10$). As shown in Table~\ref{tab:max_iteration}, surprisingly, most predictions are gotten in 1-4 iterations, and the average number of iterations is 2.43. Only a tiny portion ($\sim 0.1\%$) require the maximum number of iterations demonstrating the efficiency of the proposed approach.



\newpage
\section{More Decoding Examples}
We present more examples from the proposed Levenshtein Transformer as follows.

\begin{figure}[htpb]
    \centering
    \includegraphics[width=\textwidth]{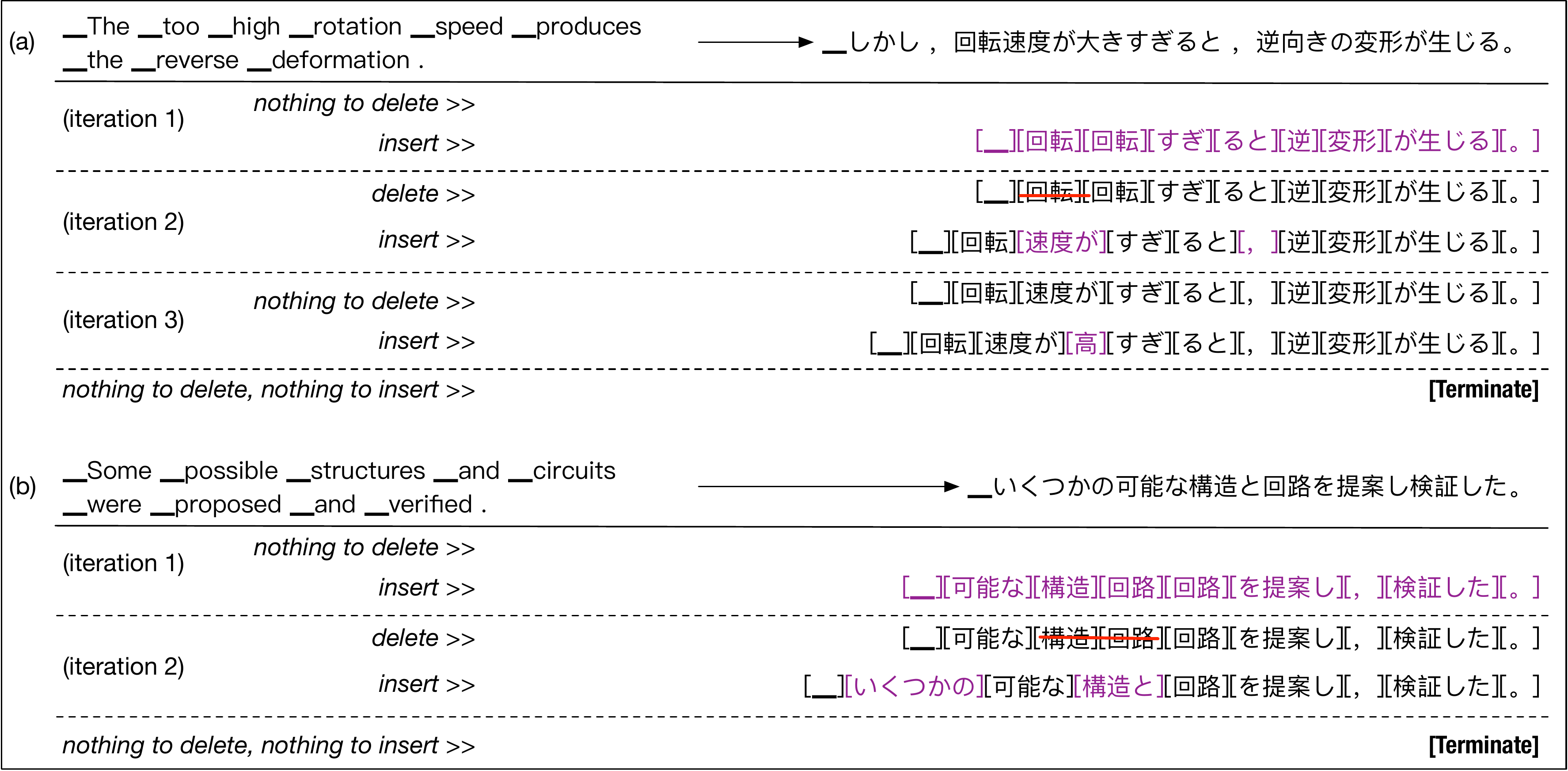}
    \caption{Translation examples for WAT'17 Small-NMT En-Ja with the Levenshtein Transformer.}
    \label{fig:example-enja}
\end{figure}
\begin{figure}[htpb]
    \centering
    \includegraphics[width=\textwidth]{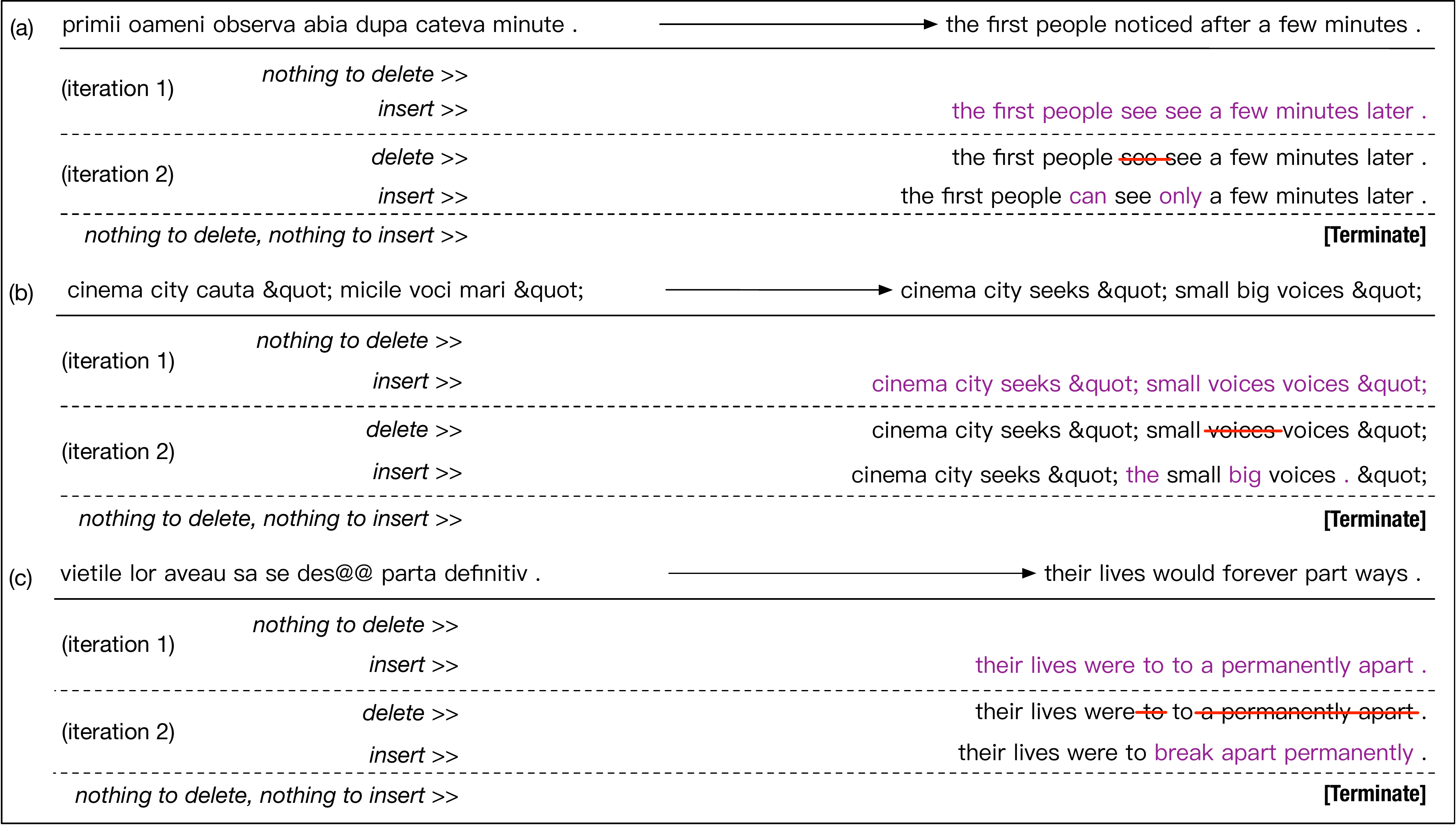}
    \caption{Translation examples for WMT'16 Ro-En with the Levenshtein Transformer.}
    \label{fig:example-roen}
\end{figure}
\begin{figure}[htpb]
    \centering
    \includegraphics[width=\textwidth]{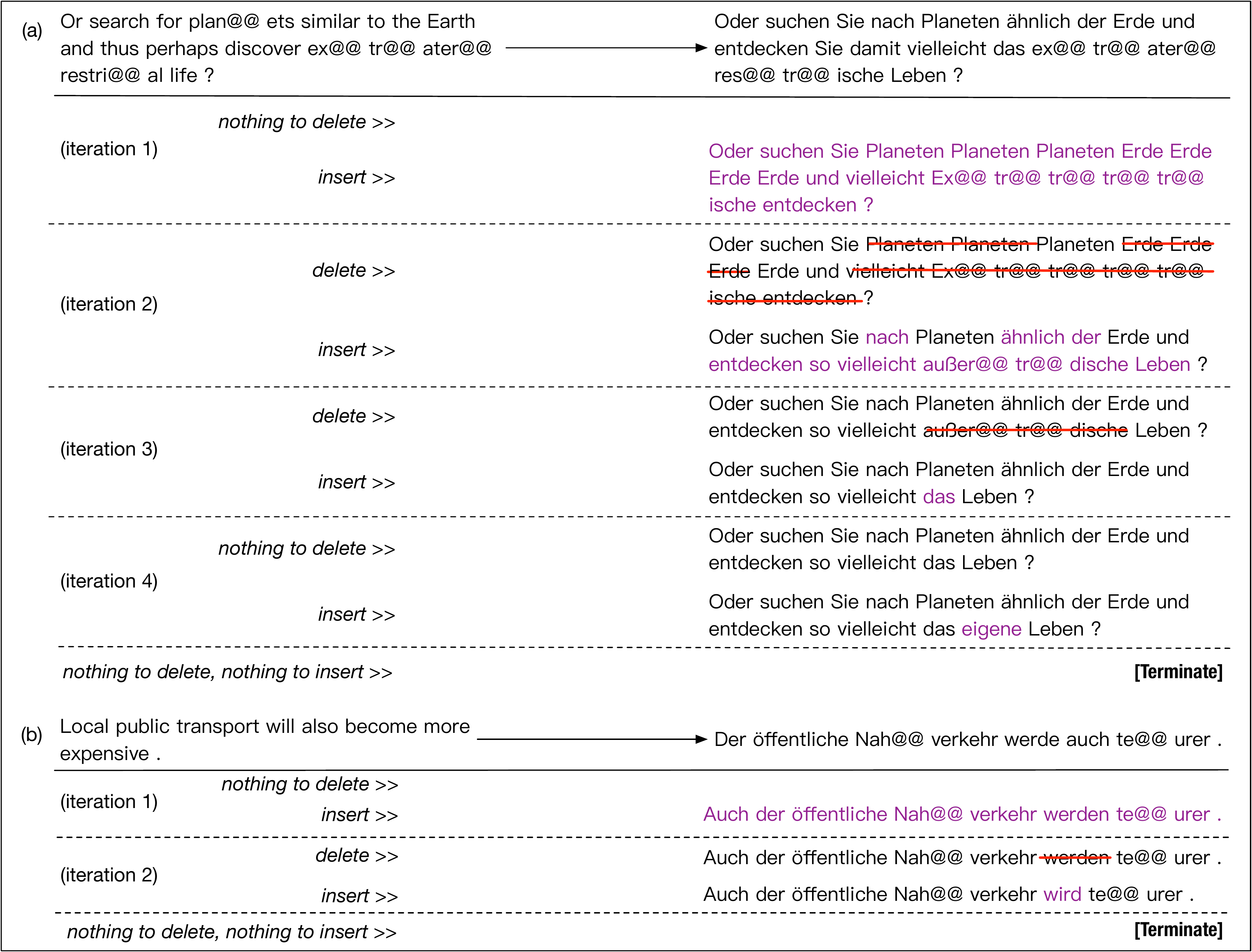}
    \caption{Translation examples for WMT'14 En-De with the Levenshtein Transformer.}
    \label{fig:example-ende}
\end{figure}
\begin{figure}[htpb]
    \centering
    \includegraphics[width=\textwidth]{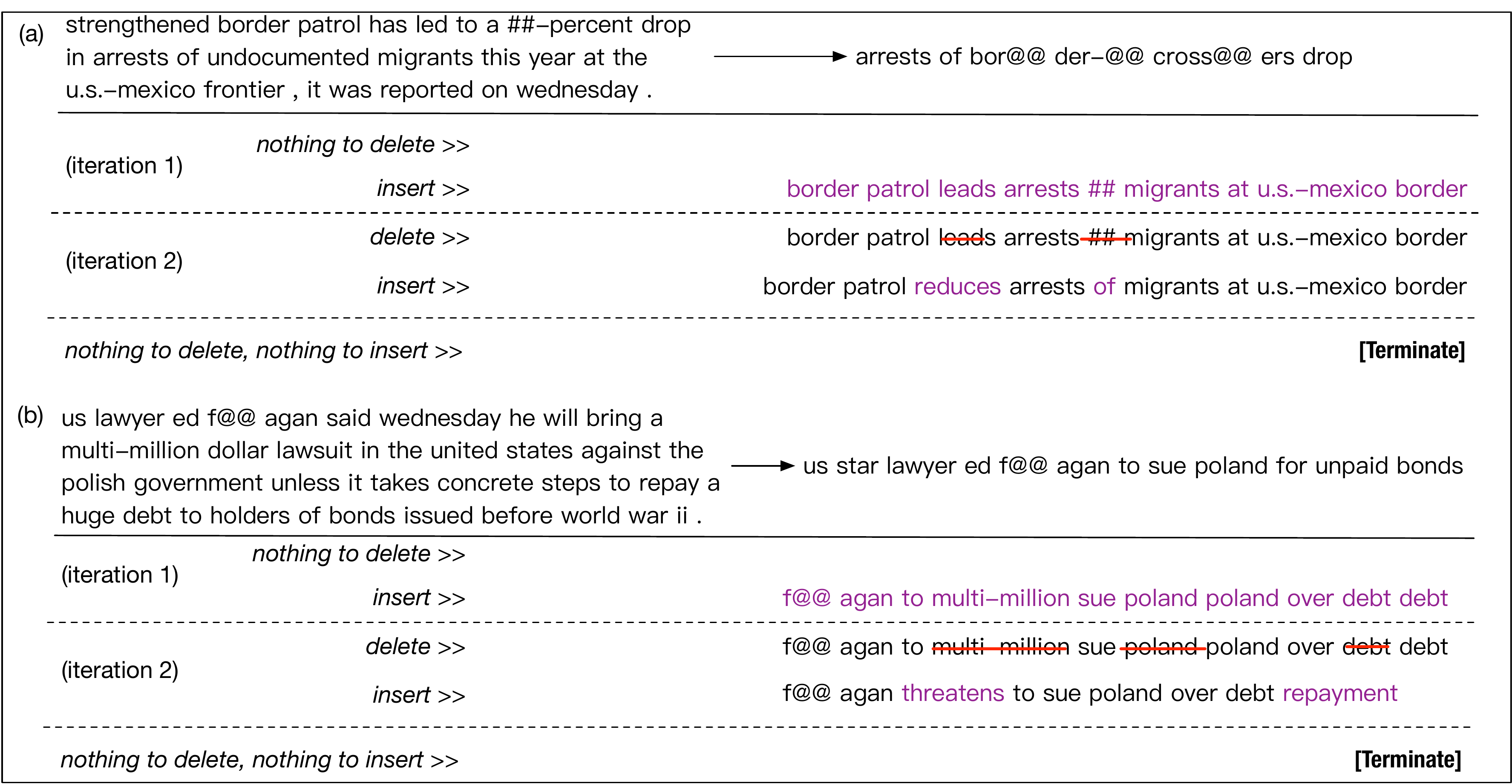}
    \caption{Translation examples for English Gigaword with the Levenshtein Transformer.}
    \label{fig:example-giga}
\end{figure}
\begin{figure}[htpb]
    \centering
    \includegraphics[width=\textwidth]{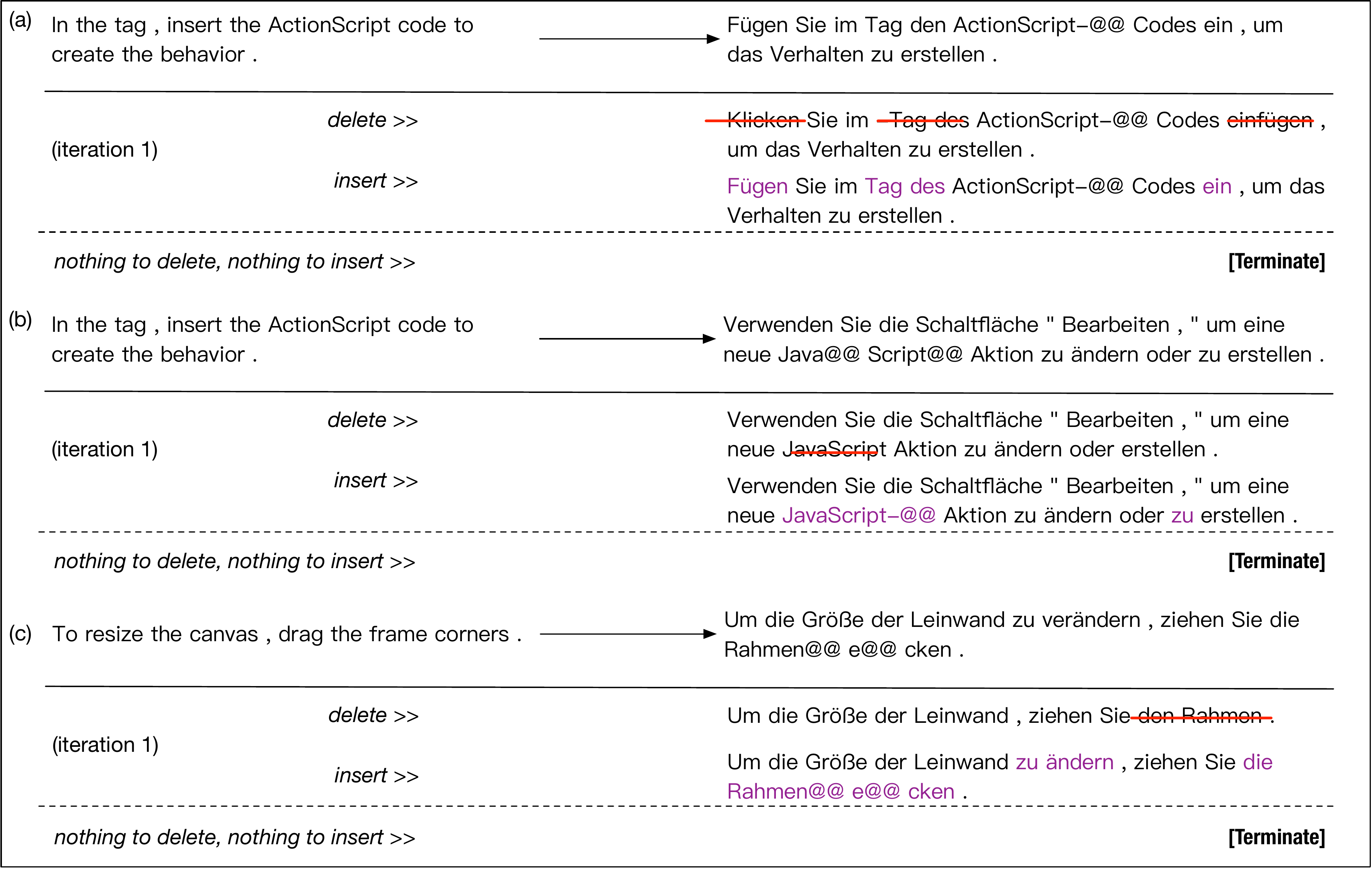}
    \caption{Post-editing examples for WMT'17-APE En-De with the Levenshtein Transformer.}
    \label{fig:example-ape}
\end{figure}
\begin{figure}[htpb]
    \centering
    \includegraphics[width=\textwidth]{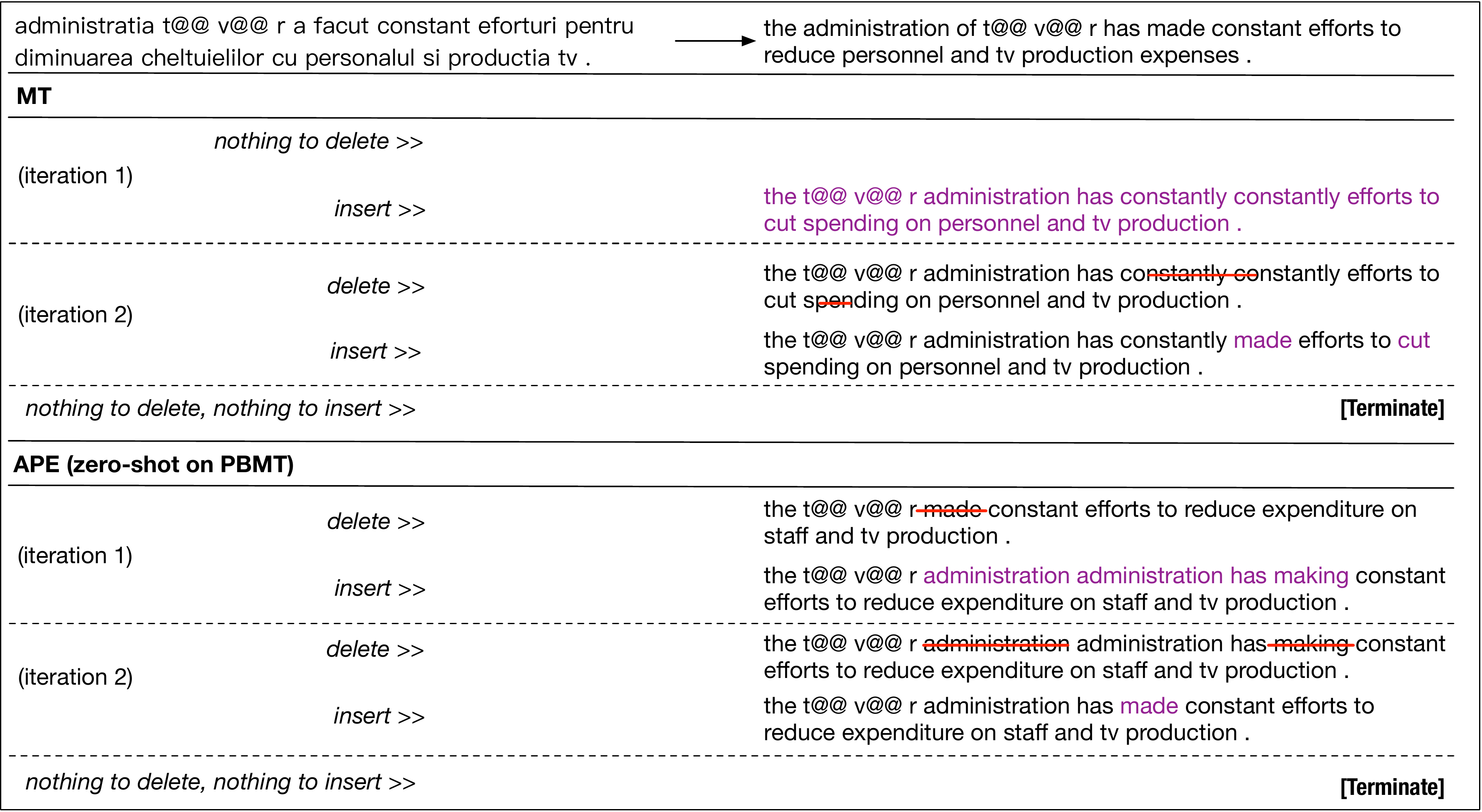}
    \caption{An example for machine translation and zero-shot post-editing over a PBMT system's output on WMT'16 Ro-En with the Levenshtein Transformer (LevT) trained for MT. It is clear to find that, the pre-trained LevT can directly adapt to the PBMT's output and have a different refinement results compared to translate from scratch.}
    \label{fig:example-mtape}
\end{figure}
\end{document}